%% file: main.tex
\documentclass[letterpaper, 10 pt, conference]{ieeeconf}  

\IEEEoverridecommandlockouts                              

\usepackage{eso-pic}
\AddToShipoutPictureBG{
 \AtPageLowerLeft{%
   \raisebox{27pt}{\makebox[\paperwidth]{\begin{minipage}{21cm}\centering
{\normalsize\color{gray}To appear at the 2026 IEEE International Conference on Robotics and Automation (ICRA). Preprint version.}
   \end{minipage}}}%
 }
}

\input{preamble}

\usepackage{comment}
\usepackage{graphicx}
\usepackage{hyperref}
\usepackage{xcolor}
\let\labelindent\relax
\usepackage{enumitem}
\usepackage[noend]{algpseudocode}
\algrenewcommand\algorithmicindent{0.6em}%
\usepackage[capitalize,noabbrev]{cleveref}
\usepackage{pgfplots}
\pgfplotsset{compat=1.18}
\usepackage{tikz}
\usetikzlibrary{fit}
\usepackage{xspace}
\usepackage{orcidlink}

\crefname{line}{line}{lines}
\crefname{figure}{Fig.}{Figs.}
\Crefname{figure}{Fig.}{Figs.}
\crefname{equation}{Eq.}{Eqs.}
\Crefname{equation}{Eq.}{Eqs.}
\crefname{section}{Sec.}{Secs.}
\Crefname{section}{Sec.}{Secs.}
\crefname{definition}{Def.}{Defs.}
\Crefname{definition}{Def.}{Defs.}
\crefname{algorithm}{Alg.}{Algs.}
\Crefname{algorithm}{Alg.}{Algs.}
\Crefname{algocf}{Alg.}{Algs.}
\Crefname{appendix}{Appendix}{Appendices}

\newtheorem{example}{Example}
\newtheorem{lemma}{Lemma}

\newtheorem{definition}{Definition}

\newtheorem{proposition}{Proposition}
\newtheorem{observation}{Observation}

\newcommand{\aref}[1]{\hyperref[#1]{Appendix~\ref*{#1}}}

\newcommand{\KE}[1]{{\textcolor{blue}{[KE: #1]}}}

\newcommand{\KEnote}[1]{}

\begin{document}
\thispagestyle{empty}
\pagestyle{empty}

\input{title}
\input{intro}

\input{prelims}

\input{correct-explanations}

\input{efficient-explanations}

\input{results}
\input{conclusion}


\bibliographystyle{IEEEtran}
\bibliography{all,more,references}

\end{document}

%% file: preamble.tex
\usepackage[utf8]{inputenc}
\usepackage{pgfplots}
\DeclareUnicodeCharacter{2212}{−}
\usepgfplotslibrary{groupplots,dateplot}
\usetikzlibrary{patterns,shapes.arrows}
\pgfplotsset{compat=newest}

\usepackage{xspace}
\usepackage{amsmath}
\usepackage{mathtools}
\usepackage{nicefrac}
\usepackage{amsfonts}
\usepackage{pifont}
\usepackage{floatflt}
\usepackage{url}
\usepackage{dsfont}

\usepackage{tikz}
\usetikzlibrary{calc, positioning, shadows, backgrounds, arrows.meta}
\usepackage{multirow}
\usepackage{booktabs}
\usepackage{tabularx}
\newcolumntype{L}{>{\raggedright\arraybackslash}X}

\usepackage{algorithm}

\makeatletter
\DeclareRobustCommand*\cal{\@fontswitch\relax\mathcal}
\makeatother

\newcommand{\xai}{\textsc{XAI}\xspace}

\newcommand{\commentout}[1]{}

\newcommand{\U}{{\cal U}}

\newcommand{\cF}{{\cal F}}
\newcommand{\V}{{\cal V}}

\newcommand{\sat}{\models}

\newcommand{\lem}{\begin{lemma}}
\newcommand{\elem}{\end{lemma}}
\newcommand{\pro}{\begin{proposition}}
\newcommand{\epro}{\end{proposition}}

\newcommand{\dfn}{\begin{definition}}
\newcommand{\edfn}{\end{definition}}

\newcommand{\xam}{\begin{example}}
\newcommand{\exam}{\end{example}}

\usepackage{algorithm}

\usepackage{longtable}
\usepackage{subcaption}
\usepackage{pgfplots}
\usepackage{pgfplotstable}
\usepackage{paralist}
\usepgfplotslibrary{fillbetween}
\pgfplotsset{compat=1.16}

\newcommand{\status}{\mathtt{status}}

\newcommand{\state}{\xi}
\newcommand{\statetraj}{\overline{\state}}
\newcommand{\statespace}{\Xi}

\newcommand{\env}{\textit{env}}

\newcommand{\obs}{z}

\newcommand{\obstraj}{\overline{\obs}}

%% file: title.tex
\title{Explaining Failures of 
Cyber-Physical Systems with Actual Causality}

\author{
Khen Elimelech$^{*,1}$\orcidlink{0000-0003-1391-8956}, Tom Yaacov$^{*,1}$\orcidlink{0000-0002-0565-6506}, David A. Kelly$^{1}$\orcidlink{0000-0002-5368-6769}, Hana Chockler$^{1}$\orcidlink{0000-0003-1219-0713}, and Moshe Y. Vardi$^{2}$\orcidlink{0000-0002-0661-5773} 
\thanks{$^{1}$ King's College London,
        {\tt\footnotesize \{firstname.lastname\}@kcl.ac.uk}}%
\thanks{$^{2}$ Rice University,
        {\tt\footnotesize vardi@rice.edu}}%
\thanks{* Equal contribution.}
\thanks{** This work was partially supported by the Causality in Healthcare AI (CHAI) Hub [UKRI AI and EPSRC grant EP/Y028856/1].}
}

\maketitle

\begin{abstract}
Modern autonomous Cyber-Physical Systems (CPSs), such as self-driving cars, face increasingly complex demands, and yet are expected to act reliably. The black-box nature often characterizing such systems, especially those relying on neural components, makes it impossible to fully verify the system behavior prior to deployment. Unfortunately, unexpected failures---cases when the system does not comply with its specification---are inevitable and may have catastrophic implications. To improve trust in the system and facilitate future mitigation after a failure occurs, it is important to try to derive an explanation for the unexpected system behavior. This paper introduces the novel concept of leveraging the framework of \emph{actual causality} for CPS failure explanation. Up~until now, this framework was only used to derive explanations in the context of simple systems, such as image classifiers. This paper addresses the theoretical gaps and provides the guidance needed to allow for correct explanation derivation in the CPS domain. Beyond the theoretical contribution, the paper presents two novel, practical, system-agnostic explanation derivation algorithms, allowing to prioritize either explanation optimality or derivation efficiency. The approach is demonstrated and evaluated in the context of a neural-network-controlled autonomous car, designed to avoid collisions.
\end{abstract}

%% file: intro.tex
\section{Introduction}
\subsection{Background: Explaining Failures}
Cyber-Physical Systems (CPSs) rely on algorithms and digital computation to control real-world machines, coordinating their components to demonstrate desired behaviors~\cite{cps,Wete2025Streamlined}. Modern CPSs, such as self-driving cars or assistive robots, are expected to achieve complex goals autonomously, i.e., without human involvement, and robustly, i.e., without failure, in a variety of complex, previously unseen environments.
With the growing demands such systems have to face, their internals also tend to grow in complexity---with modern systems almost ubiquitously leveraging ``black-box'' Deep Neural-Network (NN)-based components, such as controllers and perceptions modules, to meet these demands. The complexity and opaqueness of these systems, together with intractable environment variability, often make it that systems can only be tuned empirically, making it impossible to fully guarantee their behavior ahead of deployment. 

Unfortunately, failure cases, i.e., when the system does not comply with its given specification, are inevitable, and \emph{unexpected} failures can lead to catastrophic implications. 
Therefore, when a system is deployed and a failure is encountered, it is important to derive an \emph{explanation} for the unexpected system behavior, to help us to avoid, anticipate, or better mitigate similar failures in the future.
This process may also take place \emph{proactively}, prior to the system deployment, as part of its formal verification process; in that case, we may first smartly search for failure scenarios in simulation---a problem known as \emph{falsification}~\cite{Corso2021SurveyAlgorithms}---and, if and when those are found, derive their underlying explanations. Explaining the reasons for failure allows us to better understand system vulnerabilities and provide (some) guarantees for its behavior, improve trust in the system, and provide guidelines for system improvements. The importance of finding explanations is even more significant when the system components are well-tuned (in the NN case, -trained), making failures rarer and less expected.

Understanding the reasons for events and deriving explanations can be done through causal analysis~\cite{HP05a}.
We should note that the prevalent framework for causality nowadays, which the reader is more likely to be familiar with, is that of \emph{general} or \emph{type causality}~\cite{Pearl2009causality}. This framework is concerned with deriving forward-looking, often probabilistic, causal models, based on statistical analysis of prior experiences, for the purpose of prediction. This framework is not suitable for our objective of explaining specific (and, possibly, rare) failures.
In contrast, the study of \emph{actual causality}~\cite{Hal19} is backward-looking, and can help us build causal models and derive causal explanations underlying specific events in the past---in alignment with our interests. The framework is of aid when there are numerous potential causes for an event, whose involvement in the outcome needs to be established---as is often the case in CPS, where multiple environmental factors may affect the system's decisions.
\textbf{The high-level goal of this work is to deploy the framework of actual causality for the purpose of deriving causal explanations of failure events of (black-box) autonomous CPSs}.

\subsection{Contribution}
This paper's contribution is first and foremost \textbf{conceptual: we introduce the idea of explaining CPS failures through the lens of actual causality}.
To support this, the main objective and contribution of this paper is \textbf{applying the framework of actual causality to construct explanations of CPS behaviors and failures, and extending it as necessary}. The framework thus far has only been practically applied in the context of black-box ``classifier'' models (e.g., in~\cite{CH24, CKKS24}), and not in temporally-extended models, as in the CPS case. We start by singling out the differences between the standard ``classifier'' case and the CPS and demonstrating why naive attempts to use the framework, without addressing these differences, may lead to the derivation of incorrect explanations. We then provide the correct guidelines for explanation derivation in the CPS case.
As an additional contribution, we provide two \textbf{practical explanation-derivation algorithms} applicable for black-box CPSs.
We start by formulating a straightforward explanation algorithm that exhaustively searches the space of potential explanations and is guaranteed to return an optimal one.
While that algorithm is correct-by-construction, it is not computationally efficient. It involves searching in a high-dimensional space and running multiple system-simulation runs in perturbed, contingent environments; this process is intractable in complex scenarios.
Thus, we follow with a computationally-efficient, approximated explanation algorithm, based on a \emph{causal-responsibility}-based heuristic search. While slightly sacrificing optimality in some cases, its complexity scales more moderately with the problem size.
We conclude the paper with experimental evaluation of our contributions for an autonomous car  operating on an obstructed track.
We use this system as a running example throughout this paper, for a grounded and lucid discussion.

The paper is structured as follows:
\cref{sec:prelim} provides preliminaries on the CPS testing model and actual causality, introduces the running example, and formulates our objectives.
\cref{sec:explain-cps} provides the theoretical contribution: extension of the actual causality framework for CPSs.
\cref{sec:explanation_algo} provides the practical contribution: the failure explanation algorithms.
\cref{sec:evaluation} provides the experimental evaluation of our algorithms for the autonomous car. 
\cref{sec:conclusion} concludes the paper.

\subsection{Related work}
Attempts to define causality go back to Aristotle, with the modern view dating back to Hume~\cite{Hume39}. A more recent direction is 
\emph{but-for causality}, introduced by Lewis~\cite{Lew73}, saying that \emph{$A$ is a cause of $B$ if both happened, and if $A$ had not happened, then $B$ would not have happened}. Informally, actual causality extends and generalizes but-for causality to capture the cases of \emph{preemption} and \emph{over-determination}. 
Preemption is illustrated by the classic example of Lewis, where two children, Suzy and Billy, are
throwing rocks at a bottle. Suzy's rock hits the bottle first, shattering it. Had Suzy not thrown her rock, Billy's rock would've shattered the
bottle. Actual causality captures the fact that it is Suzy's rock that is a cause of bottle shattering, but not Billy's. Over-determination can be illustrated by
an example of two arsonists dropping lit matches in a forest. If one match is sufficient to start a fire, then none of the arsonists is a but-for cause of the fire, yet each is an actual cause of it.

Actual causality~\cite{Hal19} is well-suited for providing explanations for events that occurred. It has been used in the field of explainable AI (\xai), in particular for image classification~\cite{CH24}. An actual causality-based tool \textsc{ReX}~\cite{chockler2024causal,CKK25} is used to find minimal sets of pixels that are 
\emph{sufficient} to recreate a model's original classification of the image~\cite{Kelly25}. 
All of these applications consider a depth-$2$ causal model (see~\Cref{fig:causal_model}), a standard assumption, which we also adopt here. 
Notably, like us, \cite{CKK25} also
leverage the idea of \emph{causal responsibility}~\cite{CH04} for efficient explanations, though that technique is specifically applicable to image classification.

In the context of CPSs, \cite{araujo2025causality} uses actual causality to analyze trajectories of CPSs, but does not use causal responsibility and does not directly address the question of explanations for falsifying traces.~\cite{diehl2022} learn a Bayesian causal model from simulation data from which they generate explanations. 
This is a fundamentally different approach to the one we adopt here, which does not need to learn the internals of the model and does not need to be transferable to other scenarios. 
\cite{diwakaran2017} uses perturbations on the position of obstacles in falsifying traces to analyze CPSs. The work does not provide explanations in the causal sense, but perform a sensitivity analysis, which is also referred to as ``responsibility.'' This is not the same as causal responsibility, but can still be used to understand which parts of the environment are more strongly associated with a failing trace.
\cite{bartocci} uses a debugger to generate explanations for CPS failure. Unlike our method, which is purely black-box, that work must instrument the CPS model. This may not always be possible, especially if the model is proprietary. \cite{banerjee2021}, like our approach, examines input-output pairs, but rather than performing a causal analysis, analyses them using a (simplified) finite-state-machine version of the CPS model. Notably, such a state machine may not be available, or itself be very complex.

%% file: prelims.tex
\begin{figure}[t]
\centering
\includegraphics[width=\columnwidth]{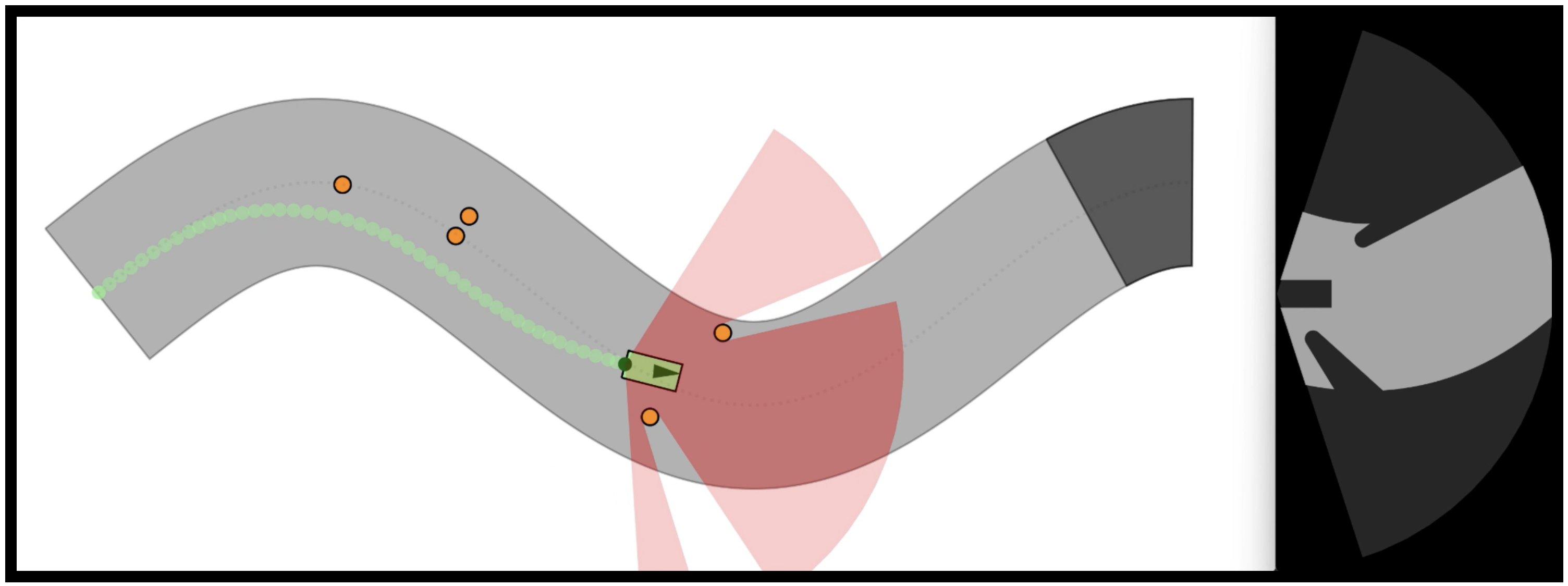}
\caption{Our running example (as introduced in~\cite{elimelech2024falsification}): an autonomous car operating in an ``obstructed track'' environment. The car trajectory is in green. At each state, the car observes the track using a lidar sensor; the observation image of the area highlighted in red is shown on the right. The car neural controller is trained to steer the car to the end of the track while avoiding collision.}
\label{fig:car}
\vspace{-10pt}
\end{figure}

\section{Preliminaries}\label{sec:prelim}

\subsection{Modeling, Testing, and Falsifying Cyber-Physical Systems}
To facilitate explanation of failures of autonomous CPSs, we adopt the formalism introduced in~\cite{elimelech2024falsification}.

\subsubsection{Environment formalism}
An \emph{environment} specifies the set of variables needed to describe the surroundings in which the system operates. An~environment can be observed by the system and potentially changed by it.
In general, variables can be classified as either \emph{parameters} or (collections of) \emph{elements}. Simply put, parameters describe (i)~essential information, which (ii)~affects the system's observation and/or dynamics globally (e.g., gravity parameters or definition of the operation area); elements describe (i)~optional, additive features used to enrich the environment, which (ii)~may be observed locally, from only a subset of system states (e.g., placement of other agents or local obstructions in the scene).

\subsubsection{The CPS model} 
We assume the system operates in an environment under continuous dynamics, modeled as:
\begin{equation}
    \begin{split}
        \dot{\state} &= f(\state,u,\env), \\
        u &= g(\obs), \\
        \obs &= h(\state,\env),
    \end{split}
\end{equation}
where $\state$~is the system state, $\env$~is the environment, $u$~is a control action, $f$~is the vector field, $g$~is a ``black-box'' controller, $z$~is the sensor observation, and $h$~is the observation (sensor) model. 
We use $\state_t$ to mark the system state at time $t\in[0,T]$, and $\statetraj_{0:T}$ to mark the concatenation of states from time~$0$ to time~$T$, a time-parameterized continuous trajectory. 
We assume the environment variability is the only source of trajectory variability, with no external uncertainty sources. 
To simplify the discussion, we ignore the initial state, and consider it to be set and predetermined.

\subsubsection{The testing model}
The testing model wraps the CPS model, as visualized in \cref{fig:testing-model}.
In this model, the simulation and test input are defined by an environment $\env$ and an initial system state $\state_0$, which together comprise a \emph{scene}.\linebreak The simulation output is the system trajectory $\statetraj_{0:T}$ (alongside $\obstraj_{0:T}$, the sensor observation history), which is used to generate the test outcome. 
We consider the controller was designed (trained) to guide the system towards completing a \emph{task}, while being robust to environment perturbation.
To evaluate the system's success in the task, i.e., the test outcome, we assume the availability of a ``status'' predicate:
\begin{equation}
    \status(\statetraj,\env) \in \{0,1\}.
\end{equation}
For a trajectory $\statetraj$, the outcome of a system run in an environment $\env$, $\status$ returns $1$, if the trajectory satisfies the task specification, or $0$, if it does not.

Including explicitly the observation history as a simulation output was shown to be useful for efficient testing. As we shall see, including the observation becomes \emph{essential}, when trying to explain a test outcome---further supporting our choice of this CPS testing model.

\subsubsection{The falsification problem}
The goal in falsification is to find (most efficiently) and return a witness of \emph{failure event}, i.e., an example of an environment $\env$ (input) and a system trajectory in it (output), for which the task specification is violated, if such a witness exists.

\subsection{A running example: an autonomous car}
\label{sec:running-example}
While the ideas presented here are relevant to a general CPS, to ground the discussion, we consider a running example of ``an autonomous car on an obstructed track,'' as depicted in \cref{fig:car}.
For this environment, the variables are: track curve, track range, track width, and a collection of obstacles. The first three are environment parameters, while the fourth component is a set of ``circular obstacle" elements, each of which of type $[x,y,r]\in\mathbb{R}^2\times\mathbb{R}^+$. 
The state $\state\doteq[x,y,\phi,\alpha,v]^T\in\statespace$ defines the car's origin position and heading, its steering angle, and speed. The observation model $h$ returns a lidar scan of the track from the car's pose (an image). The controller $g$ sets the linear acceleration and steering velocity, given the stream of lidar scans (and the current steering angle), and $f$ complies to a bicycle model.
The task for which the controller was trained is to steer the car to the end of the track while avoiding collision, regardless of the track curve and obstacle placement. 

The goal when falsifying this system is to efficiently find an environment in which running the autonomous car would result in collision.
Our goal in this work is to explain why a given falsifying environment caused the car to fail.

\begin{figure*}[t]
\centering
\begin{subfigure}{0.55\textwidth}
    \centering
    \includegraphics[width=0.9\linewidth]{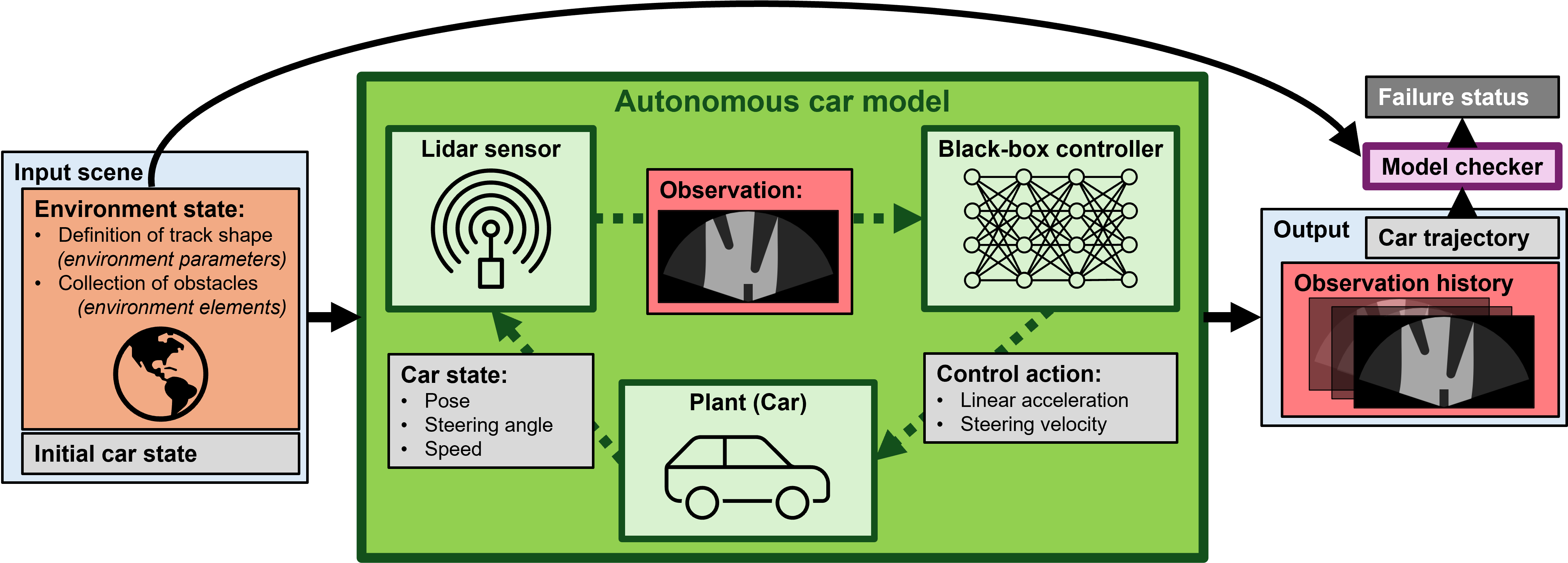}
    \caption{The CPS testing model, as formulated in~\cite{elimelech2024falsification} for falsification\\of an autonomous car system.}
    \label{fig:testing-model}
\end{subfigure}
\begin{subfigure}{0.4\textwidth}
    \centering
    \begin{tikzpicture}[outer sep=auto, scale=0.85, transform shape]
        \node (V) at (-2, 0) {$\vec{V}$};
        \node (v1) at (-1, 0) [draw, circle, minimum size=0.9cm,fill=red!10] {$v_1$};
        \node (v2) at (1.5, 0) [draw, circle, minimum size=0.9cm,fill=red!10] {$v_2$};
        \node (dots) at (2.8, 0) {$\mathbf{\cdots}$};
        \node (vn) at (4.3, 0) [draw, circle, minimum size=0.9cm,fill=red!10] {$v_n$};

        \node (f) at (1.5, -1.3)
            [draw, rectangle, rounded corners, fill=blue!10, inner sep=0.25cm]
            {CPS $\rightarrow$ Crash};

        \draw [-Triangle] (v1) -- (f);
        \draw [-Triangle] (v2) -- (f);
        \draw [-Triangle] (vn) -- (f);

        \node (o) at (1.5, -2.5)
            [draw, circle, minimum size=0.5cm, fill=yellow!10] {$O$};
        \node (oin) at (3, -2.5) {$O \in \{0,1\}$};

        \draw [-Triangle] (f) -- (o);
    \end{tikzpicture}
    \caption{A binary CPS failure-explanation causal model, with obstacle masking variables as input layer, and an indicator for crash-type-preservation as outcome.}
    \label{fig:causal_model}
\end{subfigure}
\caption{The CPS testing model and the corresponding causal model for the CPS simulation.}
\vspace{-15pt}
\end{figure*}
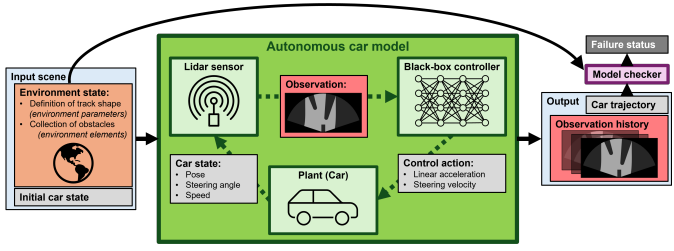
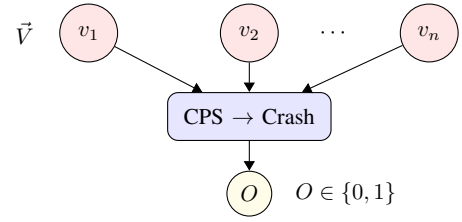

\subsection{Actual causality}\label{subsec:cause} 
In what follows, we briefly introduce the relevant concepts from the theory of actual causality. 
The reader is referred to~\cite{Hal19} for a more in-depth overview.

We assume that the world is described in terms of variables and their values.  
Some variables may have a causal influence on others.
It is useful to split the variables into two sets: the {\em exogenous\/} variables~$\U$, whose values are determined externally, and the {\em endogenous\/} variables~$\V$, whose values are ultimately determined by the values of the exogenous variables, according to a set of {\em structural equations} $\cF$.
A~\emph{causal~model}~$M$ is described by variables, their domains of values, and structural equations. 
A \emph{context}, $\vec{u}$, is the setting of a valuation for the exogenous variables $\U$, which is sufficient to determine the values of all other variables.
When all variables are binary, the model is referred to as ``binary.''

We call a pair $(M,\vec{u})$ consisting of a causal model $M$ and a context $\vec{u}$, a \emph{(causal) setting}. 
To perform causal analysis, we need to perform various modifications to some of the variables and examine their propagating effects. These modifications are formally called \emph{interventions}. An intervention is defined as setting the value of a variable $X$ to $x$, 
 denoted as $[X \leftarrow x]$. 
A Boolean formula $\varphi$ over the set of variables of $M$ may evaluate to true or false for a setting. We write $(M,\vec{u}) \sat \varphi$ if $\varphi$ is true for the setting $(M,\vec{u})$.


Given a system model $\mathcal{N}$ and a \emph{specific} event $e$ (example of a system input and corresponding output), we can define a
binary causal model $M_{\mathcal{N},e}$, which wraps and encapsulates the system model, and is used to explain the event. 
This approach is adapted from~\cite{chockler2024causal}, where it was originally developed for image classifiers.
This is done as follows:
The set of endogenous variables $\V$ contains the system input and output variables, as well as an additional binary ``outcome variable'' $O$. 
The value of the system output variables is determined by the system model, which is encoded through appropriate structural equations. The value of $O$ is determined by comparing the value in of the output variables to the outputs in the original event $e$; if those agree, the outcome value is~$1$ (True), and, otherwise, $0$~(False).
The set of endogenous variables also contains a layer of binary variables $\V^{mask}\subseteq\V$ with one-to-one correspondence with the system inputs.
These essentially act as a \emph{mask}, indicating which inputs should be fed into the system, and which should not.
To achieve that, for each system input variable $v$, we assume its domain contain a ``neutral'' value $v^0$, which is used to eliminate/minimize the effect of this variable on the system. For example, for an image classifier system, where input variables represent pixel values, this would be the color ``black'', to indicate no color.
Thus, assigning $1$ to the mask variable $v^{mask}_i$, corresponding to the system input $v_i$, would then entail (through a structural equation) that $v_i$ is assigned its original value, as in the event $e$; assigning $0$ would entail that $v_i$ is assigned the neutral value $v^0_i$.
Overall, this results in a \mbox{depth-$2$} causal model (as in \Cref{fig:causal_model}), reflecting that each input variable is (causally) independent of the others. 
The exogenous variable(s) in $\U$ are used to set a context, which determines the value of all the mask variables.
For such a model we can define two basic contexts:
$\vec{u}_1$ marks the ``original context,'' which entails setting all the mask variables to $1$, and all the system inputs to their original value, as in $e$;
and 
$\vec{u}_0$ marks the ``neutral context,'' which entails setting all the mask variables to $0$, and all the system inputs to their neutral value.
With that, we define:



\dfn[Sufficient Explanation]\label{defn:simple-exp}
For a causal model matching the description above, a subset $\V^{ex}\subseteq\V^{mask}$ of the variables is a \emph{sufficient explanation}, if the following holds:
\begin{description}
\item[{\rm EX1.}] $(M,\vec{u}_0) \models [\V^{ex} = 1](O=1)$,
where $\V^{ex} = 1$ stands for assigning $1$ to each variable $v\in \V^{ex}$.
\item[{\rm EX2.}] $\V^{ex}$ is minimal, i.e., there is no strict subset of $\V^{ex}$ that satisfies EX1.
\end{description}
\edfn
Since $\V^{mask}$ is one-to-one mapped to the system input variables, this gives us the subset of those that are sufficient (when they keep their original value) for the event outcome to occur.
When there is no confusion, we call a \emph{sufficient explanation} simply an \emph{explanation}.
Also note that this is a simplified definition, where an explanation is defined in relation to a single context, the neutral one. More generally, the explanation may be defined in relation to a set of contexts---though this will not be considered here.


As \cref{defn:simple-exp} implies, exact computation of explanations is intractable~\cite{chockler2024causal}, when the number of variables is large.
Therefore, existing explanation-derivation algorithms are typically based on approximate, heuristic search. These rely the \emph{responsibility} measure to prioritize specific variable that should be included in the explanation.
This notion of responsibility, introduced in~\cite{CH04}, is translated into our setting as follows:

\dfn[Responsibility]\label{def:simple-resp}
Consider $\V' \subseteq \V$ to be a minimal necessary set of variables for $O=o$ in a specific context $u$; that is, any intervention in one of the variables in $\V'$ changes $O$, and $\V'$ is minimal.
The \emph{degree of responsibility} of $v\in \V'$ for the outcome $O=o$ is defined as
$1/|\V'|$; otherwise, if~$v\notin \V'$, its responsibility is $0$.
\edfn


In practice, the responsibility is also intractable to calculate exactly, but it can be incrementally approximated, through sampling of random events.

\textbf{Note. } To maintain succinctness while avoiding confusion, from now on, we use ``system'' to refer to the system model, and ``model'' to refer to the causal model. 

\subsection{Problem statement}

The purpose of this paper is to properly connect the CPS testing model to the actual causality framework, to allow for explanation to CPS failure events (provided through an external falsification process).
In the following sections, we will lay out the relevant theoretical considerations when applying the actual causality framework to the CPS model, perform extension of this framework as necessary, and provide algorithms to practically derive explanation.

%% file: correct-explanations.tex
\section{Extending Actual Causality to CPS Failures}
\label{sec:explain-cps}

To explain a failure event, we must first create a causal model for it.
Though, the CPS model contains several crucial differences from the simple classifier model, for which we derived causal models thus far. These differences must be addressed when building the causal model and deriving explanations---otherwise, naive application of the actual causality framework to such systems might result in incorrect explanations. In this section, we highlight these differences and introduce the necessary mitigation to allow for a correct analysis.


\subsection{The system inputs: elements vs. parameters}
In the CPS, the environment variables, both elements and parameters, can be chosen independently and arbitrarily as input to the system.
Parameters are variables with global influence on the system. They are the first to be processed when defining the environment. The elements are variables with local influence on the system, they are processed secondarily and embedded in the environment defined by the parameters. While we are free to choose the elements independently, the expression of these variables, the way they should be embedded in the environment is determined by the environment parameters. In our example, we are free to choose the position of the obstacles, these will be placed only in relation to the track.
In technical terms, we may say that we have a causal dependency of the (expression of the) elements on the environment parameters.

Since we are currently only considering depth-2 causal models, we cannot include both parameters and elements in the model's endogenous variable set $\V$---it should include only the environment elements. Further, parameters are essential, and, hence, cannot be ``neutralised,'' as we do with elements.
Thus, according to the definitions, that the parameters should be included in the set of exogenous variables $\U$.
This means that, for each event, all the environment parameters are treated as constants, ``frozen'' with their original valuation.
This also means that the parameters (with the original valuation) are an implicit part of any explanation resulting from this model.

We should note that this is actually a sensible scenario, which means our model simply explains which environment elements caused the system to fail in the environment defined using the specific environment parameters.
In fact, we might argue that this would have been the logical choice, regardless of the technical limitation (of the model depth).
Arguably, allowing for variability of the environment parameters is incorrect, as changing a parameter can be viewed as changing the \emph{type} of failure (read: outcome), which would mean the derived explanation does not actually explain the specific event. This point is further discussed later in this section.


Another thing to note is that here, unlike the classifier case, the size of the element set is not predetermined (as an image size would be). This means that the causal model corresponding to different events of the same system may express a different topology.

\KEnote{i think effectively in the causal model we must two variables for an element - its location and its expression , which depends on its location and the track properties. this is an interesting point. should we have a new term for the elements? both dependent and independent}

\subsection{The system inputs: only considering observed elements}
In the CPS case, not
all environment elements are necessarily involved in generating the trajectory. In contrast to a classifier,
a CPS is not a simple, linear system, but a temporally-extended one, defined as a recursive loop, gradually processing localized portions of the input (environment), while building localized portions of the output (trajectory).
That is, the trajectory is built \emph{incrementally}, through sequential application of a \emph{local} observation-control-motion model. 
Furthermore, a change in an element can only cause a change in the trajectory indirectly---by affecting an \emph{observation} at a certain point in time; this in turn affects the system decision, and change the trajectory suffix.
As the environment is gradually examined, by the time the system terminates (either in failure or successfully), some elements may not have been observed.
As an example, we may think of a collision occurring early along the track; the obstacles at the end of the track are not observed and are clearly not a cause for the accident, as the car has not encountered them yet. Ignoring this observation might lead to explanations that consists of obstacles that the car has not even seen.

A more accurate depiction of a CPS as a causal model would take the temporal order into account, by introducing auxiliary variables and increasing the depth of
the model. This, however, significantly increases the complexity of computing explanations. As a simple solution in this introductory work, we may still use depth-$2$ models, but restrict explanations to
subsets of the \emph{observed} environment elements (as formally defined in~\cite{elimelech2024falsification}).
\textbf{This point should be incorporated as an amendment of \cref{defn:simple-exp}, when dealing with CPS.}
We emphasize that this amendment is necessary, and we should not mitigate this issue by restricting the endogenous variables of the causal models---while the unobserved variables cannot be a part of the explanation, they can still prevent a candidate from being an explanation.

Also, note that this identification of the observed variables should be given by the CPS simulator, in agreement with the CPS model presented before.

\KEnote{the non observed should be perturbed during the search, as changing the observed variables might mean that we now observe something we haven't observed before---are we doing that?}


\KEnote{MAYBE EXTEND THE DEF OF EXPLANATION ALSO TO DISTINGUISH BETWEEN MINIMIAL/optimal AND NON OPTIMAL, but for that we need Joe's blessings... }

\KEnote{*** observed vars  and failure types are separated points. because i can have a crash with the same obstacle but see a different subset of the elemnts before it.
though it would be hard to show them separately in this case, because on the linear track it would be hard to not see an obstacle on the way to the collision obstacle (so that would also cause the collision obstacle to change).}

\begin{figure}[t]
    \centering\small
    \begin{subfigure}{\columnwidth}\centering
        {%
        \setlength{\fboxsep}{0pt}%
        \setlength{\fboxrule}{1pt}%
        \fbox{\includegraphics[width=0.9\columnwidth]{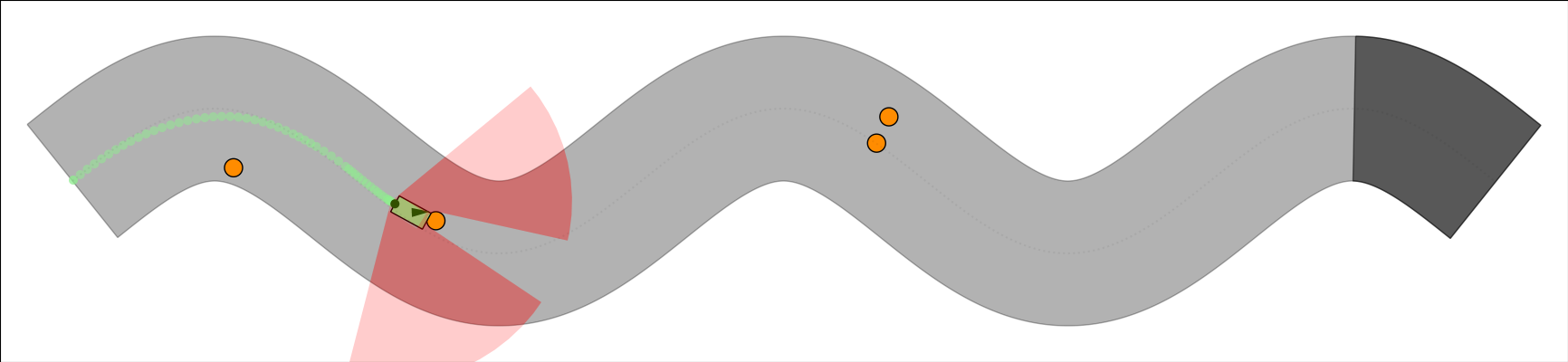} }%
        }%
    \end{subfigure}
(a)
\vspace{3pt}
    
    \begin{subfigure}{\columnwidth}\centering
        {%
        \setlength{\fboxsep}{0pt}%
        \setlength{\fboxrule}{1pt}%
        \fbox{\includegraphics[width=0.9\columnwidth]{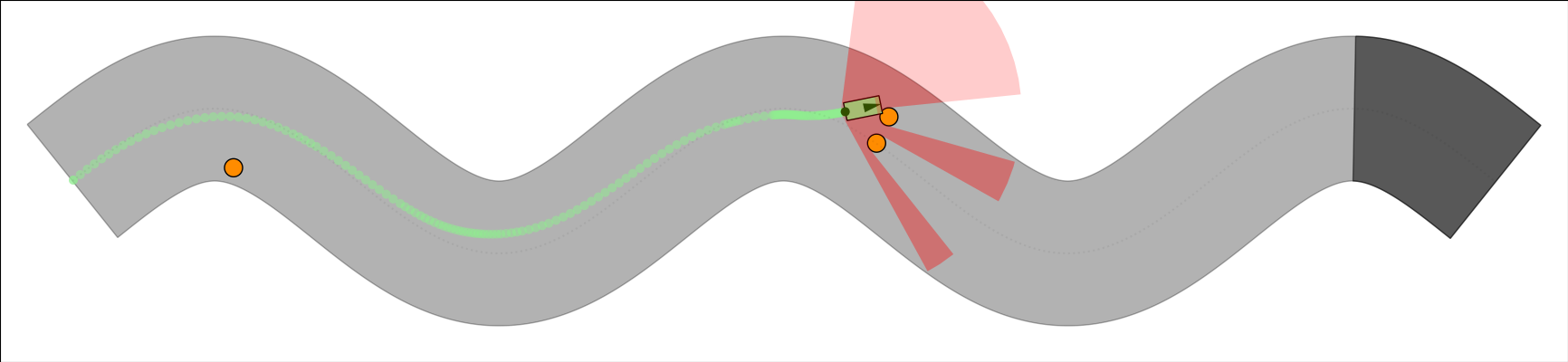} }%
        }%
    \end{subfigure}
    (b)
    \caption{Two simulation runs demonstrate our insights: (1) the third and fourth obstacles in (a) were not observed, and hence should not be considered as part of the explanation; (2) after removal of the second obstacle from (a), the car still crashes, as indicated in (b); yet, this removal changed the \emph{type} of failure. 
    This change means that the subset of obstacles in (b) is not an explanation for the crash in (a).
    }
    \label{fig:two-runs}
    \vspace{-10pt}
\end{figure}

\subsection{The system output: different types of failures}
Finally, we shall focus on the system output. In the classifier case, the output was simply a class, chosen among a discrete, finite set of classes. In the CPS testing case, we have a two-level output: the system trajectory and a binary task-success indicator derived from it.

Comparison to the system output in the given event, as we recall, defines the causal-model's binary outcome variable $O$.
A naive choice would be to treat the task success (\texttt{status}) indicator as the output.
As we recall, an explanation corresponds to the minimal set of variables (in our example, obstacles) such that, as long as they keep their values, the output remains ``the same'' as in the original event.
While success in the task is defined by examining the entire trajectory, failure, due to the local-incremental CPS model, typically happens at a certain point of the trajectory, which compromises the success criterion.
This means that failures can have many distinct ``types.'' In our example, the crash can occur in different portions of the trajectory (as visualized in \cref{fig:two-runs}).
We emphasize that we are not looking for an explanation for any failure, but for the specific one that happened.
Yet, as the task success indicator hides this information, our model would not be able to distinguish between types of failure, and may lead us to an explanation that logically does not correspond to our event. 

Since we cannot consider the task success indicator as the output, one might consider the trajectory as the output.
Yet, the trajectory is defined in a \emph{continuous} space, which would make the set of outputs also continuous and infinite. \KEnote{also the inputs, but we ignore this for now}
This, generally, means that every infinitesimally small change in the input variables can lead to a change in the output, which, in turn, would indicate that no explanation exists (because the output never remain the same).

To define the output, as we learn, we must find a midway point between these two ends of the spectrum (binary output and continuous output).
This can be done by defining an abstraction over the continuous trajectory space, grouping together similar runs to count as ``the same failures,'' to serve as the output.
This abstraction should be defined by the user, according to its logic, and shall be encoded by extending the \texttt{status} indicator from a binary to a multi-valued function.
A natural abstraction in many cases can be one differentiating between failures involving different elements.
In our example, this means categorizing failures based on the obstacle with which the collision occurred.

\cref{fig:causal_model} shows a graphical representation of our overall causal model for a CPS simulation, with arrows indicating the direction of dependency between nodes.

%% file: efficient-explanations.tex
\section{Explanation Algorithms}\label{sec:explanation_algo}

In the following, we provide two algorithms for deriving a sufficient explanation of an example of a CPS failure (i.e., a failure event) $E$, based on a causal model specified according to the guidelines provided in \cref{sec:explain-cps}.
For clarification, note that in this case, the causal model is relatively simple and there is no need to create it explicitly. The presented algorithms implicitly comply with the model when searching for an explanation.

\subsection{Exhaustive search for explanations}
The first algorithm is summarized in \cref{alg:naive-explanation}.
The algorithm performs an exhaustive search over all candidate explanations (subsets of elements in the environment of~$E$), checking each one to see whether it satisfies 
\Cref{defn:simple-exp}. Remember that, according to the causal model, we only examine subsets of the \emph{observed} elements.
In practice, this is achieved by systematically intervening in $E$, removing some of the existing elements, and checking if the original outcome (class of failure) is preserved.
If we indeed examine all candidate explanations, we can guarantee to return the subset of minimal size, as required by the definition.



\begin{algorithm}[t]
\footnotesize
\caption{Explanation for a CPS failure}
\label{alg:naive-explanation}
\renewcommand{\algorithmicrequire}{\textbf{Input:}}
\begin{algorithmic}[1]
\Require{Failure event $E$ }
\State{$elements \gets \text{element\_set\_in\_the\_environment\_of\_event($E$)}$}
\State{$O \gets \text{elements\_observed}(elements)$}
\State{$\V^{exp} \gets O$}
\For{$S \in 2^{O}$}
    \State{$elements' \gets \text{apply interv. on $O$ to align element mask with $S$}$}
    \State{$E' \gets \text{run\_simulation($elements'$)}$}
    \Statex{\Comment{Check EX1 (\Cref{defn:simple-exp})}}
    \If{$\Call{SameFailureType}{E,E'}$} 
        \If{$|S| < |\V^{exp}|$} \Comment{Check EX2 (\Cref{defn:simple-exp})}
            \State{$\V^{exp} \gets S$}
        \EndIf
    \EndIf
\EndFor
\State{\Return{$\V^{exp}$}}
\end{algorithmic}
\end{algorithm}

\begin{algorithm}[t]
\footnotesize
\caption{Efficient explanation generation}
\label{alg:explanation}
\renewcommand{\algorithmicrequire}{\textbf{Input:}}
\begin{algorithmic}[1]
    \Require{Failure event $E$, Boolean flag $\textit{MINIMIZE}$, sampling budget $m$}
    \State{$elements \gets \text{element\_set\_in\_the\_environment\_of\_event($E$)}$}
    \Statex{\Comment{Responsibility-guided ranking of elements }}
    \State{$resp \gets\Call{Responsibility\_approximation}{E, elements, m}$}
    \State{$el\_sorted \gets \text{sort\_by\_responsibility}(elements,resp)$}
    \For{$i$ in $\{ 1, \dots , |el\_sorted| \}$}\Comment{Explanation candidates evaluation}
        \State{$S \gets el\_sorted[0:i] $}
        \State{$elements' \gets$}
        \Statex{$\quad\quad\quad\quad\text{apply interv. on $elements$ to align element mask with $S$}$}
        \State{$E' \gets \text{run\_simulation($elements'$)}$}
        \If{$\Call{SameFailureType}{E',E}$} 
            \State{$\V^{exp} \gets \{ e \in elements \mid e \text{ in } el\_sorted[:i] \}$}
            \If{$\textit{MINIMIZE}$}\Comment{Explanation minimization (optional)}
            \For{$S \in 2^{\V^{exp}}$}
                \State{$elements' \gets$}
                \Statex{$\quad\quad\quad\quad\text{apply interv. on $elements$ to align element mask with $S$}$}
                \State{$E' \gets \text{run\_simulation($elements'$)}$}
                \If{$\Call{SameFailureType}{E',E}$}   
                    \If{$|S| < |\V^{exp}|$}   
                        \State{$\V^{exp} \gets S$}
                    \EndIf
                \EndIf
            \EndFor
            \EndIf
            \State{\textbf{break}}
        \EndIf
    \EndFor
    \State{\Return{$\V^{exp}$}}
    \vspace{6pt}
    \Procedure{Responsibility\_approximation}{Failure event $E$, element set $elements$, sampling budget $m$}
    \State{$resp(e) \gets 0$ for every $e$ in $elements$}
    \State{$O \gets \text{elements\_observed}(E)$}
    \State{$\mathcal{I} \gets \text{sample\_m\_random\_interventions}(O, m)$}
    \State{$candidateSubsets \gets \emptyset$}
    \For{$I$ in $\mathcal{I}$}
        \State{$E_I \gets \text{run\_simulation($I$)}$}
        \Statex{\Comment{Check necessity (\cref{def:simple-resp})}}
        \If{$\lnot \Call{SameFailureType}{E,E_I}$} 
            \State{$S \gets \text{get\_intervened\_elements}(E, I)$}
            \State{$candidateSubsets \gets candidateSubsets \cup S$}       
        \EndIf
    \EndFor
    \Statex{\Comment{Check minimality (\cref{def:simple-resp})}}
    \State{$minSubsets \gets \text{exclude\_nonminimal\_subsets($candidateSubsets$)}$}
    \For{$S$ in $minSubsets$}
        \For{$e$ in $S$}
            \Statex{\Comment{Update resp. of every element in $S$ (\cref{def:simple-resp})}}
            \State{$resp(e) \gets \text{max}(resp(e), 1/|S|)$}
        \EndFor
    \EndFor
    \State{\Return{$resp$}}
    \EndProcedure


\end{algorithmic}
\end{algorithm}

\subsection{A responsibility-guided efficient algorithm}


While the previous algorithm allows us to compute the explanation exactly, the number of candidate explanations it examines (i.e., the number of simulation runs it requires) is exponential in the number of observed elements, and hence is intractable for complex scenarios. 

Thus, we describe an efficient heuristic-search algorithm, which leverages the notion of element responsibility, in order to optimize the candidate-explanation evaluation order, prioritizing explanations with high-responsibility elements. By such, this technique should reduce the number of simulation runs needed to yield an explanation, compared to the exhaustive algorithm.
This algorithm essentially invests some of its computational budget in pre-processing, in order to estimate the responsibilities, in hopes that this would lead to reduced effort of the explanation search.

We note that in the previous algorithm, the minimality of the explanation is guaranteed, since we check all element subsets.
As this is not the case here, once a candidate passes the check (preserving the class of failure), we may follow up with an attempt to minimize it---essentially, performing an exhaustive search on this specific element subset.
With this additional step, the algorithm, as we demonstrate in \cref{sec:evaluation}, in the majority of the cases returns an exact (minimal) explanation, while also reducing the number of simulation runs, compared to the baseline approach.

This algorithm, which is summarized in \cref{alg:explanation}, is divided into three main steps: (i) compute an approximation of the degree of responsibility for each element, (ii) use these measures to heuristically search for a (non-minimal) explanation; and (iii) optionally, attempt to minimize this explanation.

The ``responsibility approximation'' procedure in \cref{alg:explanation} starts by removing all elements that were not observed by the CPS from inclusion in the explanation (line 21). In line 22, it creates a dataset $\mathcal{I}$ of $m$ random CPS run examples, through random interventions over the original example $E$. In lines 24-32, the procedure uses $\mathcal{I}$ to iteratively estimate the degree of responsibility for each element, in accordance with \cref{def:simple-resp}.
%
%
After the responsibilities are calculated, the algorithm ranks the elements accordingly (line 3). Then, in lines 5-7, the algorithm incrementally evaluates prefixes of this ordering as explanation candidates, until running the simulation outputs the same class of failure as in $E$; this marks the explanation to be returned. An optional step, in lines 11-16, tries to optimize the candidate by looking for a smaller explanation in its powerset.

%% file: results.tex
\section{Experimental evaluation}
\label{sec:evaluation}
Next, we demonstrate and quantitatively compare our proposed explanation algorithms for finding explanations of CPS failures, in the context of our running example.


\textbf{Experimental Setting:} For experimentation, we used the LiteRacer~\cite{Elimelech24LiteRacer} simulator. This is an open-source Python engine for simulating an autonomous car on an obstructed track. The car was powered by the default NN-based controller provided with the simulator, and all simulation properties (car speed, sensor range, etc.) were unmodified from their default values.
With LiteRacer, we generated a collection of $32$ falsifying simulation run examples (failure events). For that, we used the state-of-the-art ``meta-planning'' falsification algorithm, developed in \cite{elimelech2024falsification} and already implemented in LiteRacer.
Each failure event corresponds to a different environment in which the car crashes.
As the track shape (environment parameters) was kept consistent in all events, we grouped them simply according to the number of obstacles, indicating the environment complexity: events with 6, 9, 12, and 15 obstacles (8 examples of each). 
Experiments were conducted on an Xeon E5-2620 CPU with 16GB RAM.


\textbf{Comparison:} To evaluate our contribution, we derived explanations for each failure event, in three ways\footnote{Implementation of the algorithms is available at~\url{https://github.com/AutonomousRobotsLab/LiteRacer_explanations}.}:
\begin{enumerate}
    \item Exhaustive Search (ES): as suggested in~\cref{alg:naive-explanation}.
    \item Responsibility-Guided (RG): as suggested in~\cref{alg:explanation}, without the minimization step ($\textit{MINIMIZE}=$ \emph{false}).
    \item Responsibility Guided with Minimization (RGM): as suggested in~\cref{alg:explanation}, with the minimization step ($\textit{MINIMIZE}=$ \emph{true}).
\end{enumerate}
We recall that the RG and RGM algorithms invest some of their computational budget in preprocessing, in an attempt to reduce the effort of the explanation search.
Thus, for each RG and RGM, we considered three sub-variants, in which we used different computational budgets (that is, the number of sampled simulation runs allocated) to approximate the responsibility: using 100, 200, and 300 samples.
For each approach, we evaluated the computational effort it required to find explanations (in terms of the total number of simulation runs), and the accuracy of those explanations, in terms of the explanation size (i.e., the number of elements that make up the explanation).
The ES approach served as the baseline for evaluating our approach, in terms of both effort and accuracy, as it is guaranteed to yield a minimal-size explanation.

\begin{table}[t]
\centering
\caption{Median explanation size yielded by the algorithmic variants, expressing solution accuracy. Lower is better.}
\begin{tabular}{l|| r *{6}{r}}

\multicolumn{2}{c}{} & \multicolumn{2}{c}{\bfseries 100 samples } & \multicolumn{2}{c}{\bfseries 200 samples } & \multicolumn{2}{c}{\bfseries 300 samples }  \\
\cmidrule(lr){3-4}\cmidrule(lr){5-6}\cmidrule(lr){7-8}\
 \bfseries \#Obs & \bfseries ES &
\bfseries RG & \bfseries RGM &
\bfseries RG & \bfseries RGM &
\bfseries RG & \bfseries RGM \\
\midrule
6  & \textbf{2}  & 5  & \textbf{2} & 5  & \textbf{2} & 5  & \textbf{2}  \\
9  & \textbf{2.5}  & 5.5 & \textbf{2.5} & 6 & \textbf{2.5} & 6  & \textbf{2.5}   \\
12 & \textbf{3}  & 9 & \textbf{3} & 7 & \textbf{3} & 7  & \textbf{3}   \\
15 & \textbf{3}  & 9.5  & \textbf{3} & 7.5  & \textbf{3} & 7   & \textbf{3} \\
\bottomrule
\end{tabular}
\label{tab:explanation-size}
\end{table}

\begin{table}[t]
\centering
\setlength{\tabcolsep}{2pt}
\caption{Median explanation effort (amount of simulation runs) required by the algorithmic variants; the number in brackets indicates the search effort, without preprocessing. Lower is better; \textbf{x} signifies best value; \underline{x} signifies best value among variants that returned an optimal explanation.}
\begin{tabular}{l|| r r *{6}{r}}

\multicolumn{2}{c}{} & \multicolumn{2}{c}{\bfseries 100 samples } & \multicolumn{2}{c}{\bfseries 200 samples } & \multicolumn{2}{c}{\bfseries 300 samples } \\
\cmidrule(lr){3-4}\cmidrule(lr){5-6}\cmidrule(lr){7-8}
 \bfseries \#Obs & \bfseries ES &
\bfseries RG & \bfseries RGM &
\bfseries RG & \bfseries RGM &
\bfseries RG & \bfseries RGM \\
\midrule
6  & \underline{\textbf{64}}    & 105 (5) & 137  (37)& 205 (5) & 237 (37) & 305 (5) & 337 (37) \\
9  & 192   & \textbf{105} (5) & \underline{177}  (77)& 206 (6) & 342 (142) & 306 (6) & 442 (142) \\
12 & 1024  & \textbf{109} (9) & 621  (521)& 207 (7) & \underline{335} (135) & 307 (7) & 435 (135) \\
15 & 1152  & \textbf{109} (9) & 1261 (1161)& 207 (7) & \underline{399} (199) & 307 (7) & 435 (135)\\
\bottomrule
\end{tabular}
\label{tab:runs}\vspace{-10pt}
\end{table}

\textbf{Results:} 
\Cref{tab:explanation-size} presents the median size of the generated explanations for each of the examined algorithms and for each group of events.
As expected, RG generated larger (i.e., less exact) explanations, with the ratio compared to ES increasing with the number of obstacles. 
RGM, however, managed to find the accurate explanation in all cases---indicating the importance of the minimization step.
%
\Cref{fig:runs} and~\Cref{tab:runs} present the median number of simulation runs for each of the examined algorithms and for each group of events. The results show that RG drastically outperformed RGM and ES in terms of efficiency, remaining nearly constant across different obstacle counts and sample sizes---though this approach, as mentioned, sacrificed the explanation accuracy. RGM, nevertheless, still required less effort than ES and also maintained the solution accuracy. Further, we can identify that allocating 200 samples for preprocessing seemed to be the most cost-efficient choice; while increasing the number of samples to 300 managed to improved the search efficiency, it was not improved enough to justify the investment (though this variation still outperformed ES).

The results demonstrate a tradeoff between explanation accuracy and computational effort. ES guarantees accuracy, but requires a large amount of simulation runs, which grow exponentially with the number of obstacles observed in the event. In contrast, RG requires drastically less runs but produces explanations that are less accurate (larger), with respect to those discovered by ES.
Positioned between these extremes, RGM appears as the most balanced approach. While it is not guaranteed, RGM practically achieved the exact explanations in all examined cases, and still required a lower number of simulation runs compared to ES---scaling much more moderately to the growing number of obstacles.

%% file: conclusion.tex
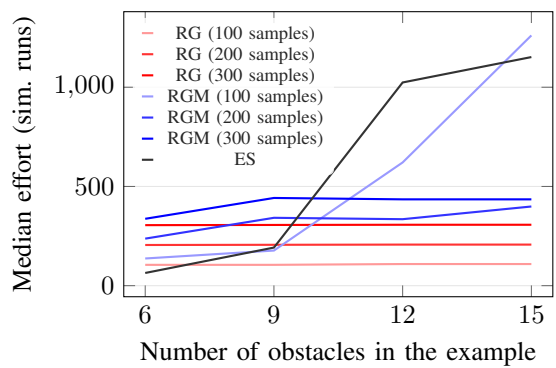
\begin{figure}[t]
    \centering
    \begin{tikzpicture}
\begin{axis}[
    width=0.84\linewidth,
    height=0.62\linewidth,
    xlabel={Number of obstacles in the example},
    ylabel={Median effort (sim. runs)},
    xmin=5.5, xmax=15.5,
    xtick={6,9,12,15},
    ymajorgrids,
    grid style={gray!20},
legend style={
  at={(rel axis cs:0.02,0.98)}, 
  anchor=north west,
  draw=none,
  fill=white, fill opacity=0.85,
  rounded corners=2pt,
  font=\scriptsize,
  row sep=0.5pt,
  column sep=4pt,
  nodes={inner sep=1pt, text depth=0.15ex}
  },
legend image post style={xscale=0.25, yscale=0.25},
    tick style={black!60},
    tick label style={/pgf/number format/fixed},
    every axis plot/.append style={thick},
]


\addplot[color=red!40]
    coordinates {(6,105) (9,105) (12,109) (15,109)};
\addlegendentry{RG (100 samples)}
\addplot[color=red!80]
    coordinates {(6,205) (9,206) (12,207) (15,207)};
\addlegendentry{RG (200 samples)}
\addplot[color=red]
    coordinates {(6,305) (9,306) (12,307) (15,307)};
\addlegendentry{RG (300 samples)}

\addplot[color=blue!40]
    coordinates {(6,137) (9,177) (12,621) (15,1261)};
\addlegendentry{RGM (100 samples)}
\addplot[color=blue!80]
    coordinates {(6,237) (9,342) (12,335) (15,399)};
\addlegendentry{RGM (200 samples)}
\addplot[color=blue]
    coordinates {(6,337) (9,442) (12,435) (15,435)};
\addlegendentry{RGM (300 samples)}

\addplot[color=black!80]
    coordinates {(6,64) (9,192) (12,1024) (15,1152)};
\addlegendentry{ES }
\end{axis}
\end{tikzpicture}
\caption{Median explanation effort required by ES, and the RG and RGM variants, per set of examples. Lower is better.}
\label{fig:runs}
\vspace{-5pt}
\end{figure}

\section{Conclusion}
\label{sec:conclusion}
This paper opened a novel direction towards better understanding and handling of failures of autonomous CPSs---one that is rooted in the rigorous mathematical theory of actual~causality. 
As this paper demonstrated, the unique properties 
of CPS lead to several non-trivial implications, which must be addressed in order to derive correct explanations. We discussed how one should build causal models in this case and followed with two practical, system-agnostic explanation-derivation algorithms, allowing to prioritize explanation accuracy or efficiency.
It is important to note that, as a direct continuation to the prior work this work builds on, we only considered here \emph{sufficient} explanations: minimal subsets of obstacles that are sufficient for the failure to occur. These explanations are useful for localizing concerns (e.g., if there are several failure-inducing regions in the environment) and for generating new failure examples. The latter is especially important in the CPS case, where failures are often rare and, hence, difficult to simulate and debug.

In the future, we plan to continue enriching this framework in two main avenues. The first will be achieved by supporting more complex causal models, e.g., to properly capture the continuous and temporally-extended nature of CPSs. 
The second will be achieved by examining other types of explanations, namely, \emph{necessary} explanations (minimal subsets of elements whose removal eliminates the failure); these are complementary to the sufficient explanations computed here and may potentially be more suitable to failure explanation.

This paper serves as a crucial step towards explainable and trustworthy autonomous CPSs and robots.